\documentclass[10pt,twocolumn,letterpaper]{article}

\usepackage{wacv}
\usepackage{times}
\usepackage{epsfig}
\usepackage{graphicx}
\usepackage{amsmath}
\usepackage{amssymb}

\usepackage{float}
\usepackage{subfig}
\usepackage{arydshln}

\newcommand{\A}{\mathcal{A}}
\renewcommand{\S}{\mathcal{S}}
\newcommand{\Sp}{\hat{\S}}
\newcommand{\Ap}{\hat{\A}}
\newcommand{\Sgt}{\S_{gt}}
\newcommand{\Agt}{\A_{gt}}
\newcommand{\I}{\mathcal{I}}
\newcommand{\Igt}{\I_{gt}}
\newcommand{\D}{\mathcal{D}}
\newcommand{\N}{\mathcal{N}_{\phi}}
\newcommand{\C}{\mathcal{C}}
\newcommand{\Cgt}{\C_{gt}}
\newcommand{\Cp}{\hat{\C}}
\renewcommand{\P}{\mathcal{P}}
\newcommand{\Pp}{\hat{\P}}
\DeclareMathOperator*{\argminx}{argmin} 
\newcommand{\argmin}[1]{\argminx_{#1}}
\newcommand{\MSE}{MSE\xspace}
\newcommand{\siMSE}{si-MSE\xspace}
\newcommand{\siLMSE}{si-LMSE\xspace}
\newcommand{\rsMSE}{rs-MSE\xspace}
\newcommand{\SSIM}{{SSIM}\xspace}
\newcommand{\DSSIM}{{DSSIM}\xspace}
\newcommand{\Lgrad}{L_{\nabla}}
\newcommand{\Ladv}{L_{adv}}
\newcommand{\nbpx}{N}
\newcommand{\nbpatches}{P}

\newcommand{\oursff}{\textcolor{magenta}{$\bigstar$} (50/50)\xspace}
\newcommand{\oursht}{\textcolor{cyan}{$\bigstar$} (union)\xspace}

\graphicspath{{imgs/}}

\wacvfinalcopy 

\def\wacvPaperID{76} 

\ifwacvfinal\fi
\begin{document}

\title{DARN: a Deep Adversarial Residual Network for Intrinsic Image Decomposition}

\author{Louis Lettry\\
CVL, ETH Z\"urich\\
{\tt\small lettryl@vision.ee.ethz.ch}
\and
Kenneth Vanhoey\\
CVL, ETH Z\"urich\\
{\tt\small kenneth@research.kvanhoey.eu}
\and
Luc van Gool\\
CVL, ETH Z\"urich\\
PSI -- ESAT, KU Leuven\\
{\tt\small vangool@vision.ee.ethz.ch}
}

\maketitle
\ifwacvfinal\thispagestyle{empty}\fi

\begin{abstract}
We present a new deep supervised learning method for intrinsic decomposition of a single image into its albedo and shading components.
Our contributions are based on a new fully convolutional neural network that estimates absolute albedo and shading jointly.
Our solution relies on a single end-to-end deep sequence of residual blocks and a perceptually-motivated metric formed by two adversarially trained discriminators.

As opposed to classical intrinsic image decomposition work, it is fully data-driven, hence does not require any physical priors like shading smoothness or albedo sparsity, nor does it rely on geometric information such as depth.
Compared to recent deep learning techniques, we simplify the architecture, making it easier to build and train, and constrain it to generate a valid and reversible decomposition.
We rediscuss and augment the set of quantitative metrics so as to account for the more challenging recovery of non scale-invariant quantities.

We train and demonstrate our architecture on the publicly available MPI Sintel dataset and its intrinsic image decomposition, show attenuated overfitting issues and discuss generalizability to other data. 
Results show that our work outperforms the state of the art deep algorithms both on the qualitative and quantitative aspect.
\end{abstract}


\section{Introduction}
The image formation process is a complex phenomenon of light entering a scene, being transformed and reaching an observer.
Barrow and Tenenbaum~\cite{BT78} define it as a mixture of the intrinsic scene characteristics like range, orientation, reflectance and illumination.
Being able to reverse the process by decomposing an image into intrinsic components is a useful pre-process for many computer vision and graphics tasks.
\begin{figure}
    \begin{center}
        \subfloat{\includegraphics[width = \linewidth]{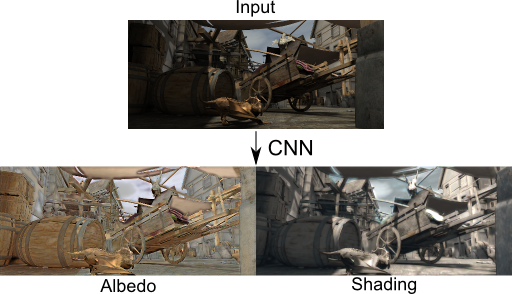}}
    \end{center}
    \vspace{-0.2cm}
    \caption{Intrinsic image decomposition separates $\I$ in a pixel-wise product of albedo $\A$ and shading $\S$, here using our contribution.
    Albedo represents the base colors of the objects while shading includes all lighting-induced effects.}
    \label{fig:N}
    \vspace{-0.4cm}
\end{figure}
We use deep learning to tackle the task of intrinsic image decomposition, which aims at splitting an image composed of diffuse materials into the per-pixel product of diffuse albedo $\A$ (\ie, base color) and shading $\S$ such that: $\I=\A\cdot\S$ (see Fig.~\ref{fig:N}).
Albedo is a lighting-invariant quantity, while shading gives important cues on the light environment and object geometry and material. 
In computer graphics, intrinsic decomposition is at the base of shadow removal, and scene relighting or recoloring~\cite{BKPB17}, which are all essential in augmented reality.
Typical vision tasks (\eg, shape from shading, 3D reconstruction, depth or normal estimation) benefit from a shading-only image, while segmentation conversely is mainly interested in the albedo image.

For many of these tasks, it is essential that the predicted decomposition is accurate.
This is a hard task due to the inherent ambiguity on scale, influenced by unknown exposure at capture time.
Therefore, the state of the art methods predict two quantities $\A$ and $\S$ regardless of scale, and both are separately evaluated on scale-invariant metrics~\cite{directintrinsic}.
The output thus requires manual tuning of scale before being usable, and there is no guarantee on energy preservation: $\I \neq \A\cdot\S$.
Finally, despite an elaborated multi-scale convolutional neural network (CNN) architecture that requires substantial training effort, qualitative results show improvement potential.

We argue that
i) predictions should be energy-preserving, and
ii) $\A$ has a unique absolute value (defined by physical properties) we should aim at.
In this context, we propose a deep generative CNN that predicts albedo and shading jointly with increased accuracy and ensuring energy preservation by design, along with new scale-sensitive evaluation metrics.

We design our network based on three key observations.
First, we conjecture that predicting $\A$ and $\S$ separately requires additional training effort and is prone to inconsistencies that violate $\I=\A\cdot\S$, hence we learn them fully jointly and interdependently.
Second, $\I$ is a good initial guess for both $\A$ and $\S$.
Hence, inspired by residual networks~\cite{resnet}, we learn a deviation from the input rather than from zero.
Third, state of the art results show typical generative CNN artifacts like blur, color bleeding and contrast oscillations.
We think they can be avoided by using a loss function in the form of a discriminator network~\cite{GAN}.
We leverage these observations into the following contributions:
\begin{itemize}\vspace{-2mm}
 \item a powerful residual generative adversarial network (GAN) for fully joint learning of $\A$ and $\S$,\vspace{-2mm}
 \item new stricter quantitative metrics that incorporate consistency and scale, and that we discuss and motivate,\vspace{-2mm}
 \item a new state of the art in intrinsic image decomposition, outperforming previous work qualitatively and quantitatively on dense ground truth data, both on scale-invariant --and sensitive metrics.
\end{itemize}\vspace{-2mm}
Data and code will be made available upon publication.

\section{Related Work}
\paragraph{Intrinsic decomposition.}
The image formation process can be defined as the pixel-wise product of the albedo $\A$ defining a surface's base color, and the shading $\S$ defining the captured influence of light reflection and shadowing on each pixel:
 $\I=\A\cdot \S$\,
(see Fig.~\ref{fig:N}).
Intrinsic decomposition is the process of estimating $\A$ and $\S$ from $\I$: see~\cite{BarronTPAMI2015} for a comprehensive recent review.
This is a hard ill-posed problem as it is highly underdetermined and requires strong and accurate priors to extract valid solutions.
Hence, prior work focuses on adding constraints to produce a deterministic plausible solution.

In the single-image case, prior statistics on both~$\A$ and~$\S$~\cite{BarronTPAMI2015,RobotVisionRetinex,Shen2008intrinsic,ZhaoRetinexTexture} and/or manual user intervention~\cite{BousseauInteractive09,ShenReflectance11} is used.
Priors are typically the \emph{retinex} assumptions~\cite{retinex}: slow variation (\ie, smoothness) of $\S$, which correlates with the normals in the scene, and piecewise-constancy of $\A$, which is generally unique per object or pattern in the scene.
This works well in a Mondrian world~\cite{Chen13}, but fails on complex or natural scenes.
\emph{Bell}~\etal~\cite{IntrinsicImageInWild14} complements the priors by deducing an albedo palette from user annotations and optimize a decomposition using conditional random fields.

Depth cues in the form of a 3D proxy or depth maps impose important local constraints on reflectance thus shading~\cite{Lee2012,Chen13}.
However, depth either has to be measured alongside the image (\eg, with a depth sensor) or deduced from multiple images with varying lighting~\cite{DucheneTOG15} or viewpoints~\cite{HachamaICCV15,LaffontSigAsia12}.
Non-explicit depth can also be exploited by sophisticated reasoning on the variation in pixel color with varying lighting~\cite{LaffontICCV15}.
Conversely, we take the image as only input.

The availability of larger datasets (see below) allowed deep learning techniques to be applied on the single-view case.
\emph{Narihira}~\etal~\cite{directintrinsic} propose a multi-scale CNN architecture composed of two branches so as to treat both global features and local details.
These branches are merged into a single one, which is later split again to predict~$\A$ and~$\S$ separately.
The resulting~$\A$ and~$\S$ are scale-invariant and not consistent: it violates $\I=\A\cdot\S$.
This can cause problems to several applications, like normal estimation or shape from shading, and may require prior manual correction.
\emph{Zhou}~\etal~\cite{ZhouICCV15} augment the retinex priors by (deep) learning from crowd-sourced sparse annotations, and inject the result into a CRF that predicts $\A$ and $\S$.
Similarly to~\cite{IntrinsicImageInWild14}, qualitative and quantitative results \wrt the sparse annotations (\ie, the WHDR metric) are good, but none are given \wrt dense annotations.
We also propose a CNN, with some major changes that guarantee energy preserving consistent predictions that outperform the state of the art on dense ground truth (GT) data.

\paragraph{Deep learning.}
The training of CNN has greatly benefited from several architectural advancements we will take advantage of.
Residual networks (ResNet) estimate the difference from the input instead of a full mapping from scratch~\cite{resnet}.
Because $\A$ (respectively $\S$) and $\I$ have a lot in common, we argue that this is beneficial to our problem.

Generative adversarial networks (GAN) append a discriminator network to a generator network~\cite{GAN}.
The role of the former is to distinguish between an artificial output and a GT (albedo or shading, in our case).
It is jointly trained with the generator, who tries to fool the discriminator by producing indistinguishable outputs.
In that sense, a discriminator can be seen as a perceptual loss function.
Exploiting this allows us to outperform prior works which suffer from typical CNN artifacts like blurriness, color bleeding and contrast oscillations.

\paragraph{Datasets.}
Data-driven approaches~\cite{LaffontICCV15,directintrinsic,ZhouICCV15} learn priors during training, either explicitly or implicitly.
GT databases can be leveraged for that.
However, acquiring GT albedo and shading images in the real world is difficult: it requires a precise controlled white and uniform light environment, hence is not scalable.

The MIT intrinsic images dataset focuses on single (segmented) objects~\cite{MITintrinsicImages}.
Due to the small number of elements in the dataset (\ie, 20 objects), training a CNN on the MIT set alone would inevitably result in a low generalization capacity.
\emph{Laffont}~\etal~\cite{LaffontICCV15} similarly focus on a single synthetic outdoor building under multiple viewpoints and lighting conditions.
Conversely, \emph{Intrinsic images in the wild}~\cite{IntrinsicImageInWild14} provide sparse relative user annotations on indoor scenes.
Finally, the Sintel short animation movie has been made publicly available with its different rendering layers, including albedo and shading~\cite{SINTELdataset}.
It forms the only large dense GT dataset, but is biased towards unnatural colors (\eg, blue and gold colors and fluorescence).
We train (section~\ref{sec:training}) and evaluate (section~\ref{sec:results:sintel}) on it, and study generalizability to other data domains in section~\ref{sec:results:generalization}.

\section{Deep Intrinsic Decomposition}
\label{sec:intrinsicdecomposition}
We decompose an image $\I(x) \in \mathbb{R}^3$ into albedo~$\A(x) \in \mathbb{R}^3$ and shading $\S(x)\in \mathbb{R}^3$ following
\begin{equation}
\label{eq:original_eq}
	\I(x) = \A(x) \cdot \S(x)\;,
\end{equation}
where $ \cdot $ denotes the element-wise product, and $x$ a pixel (that we will drop in all following notations).
Note that shading has 3 channels to allow for colored illumination.

The decomposition problem is ill-posed:
many possibilities for~$\A$ and~$\S$ respect the above equation, and the scale of $\I$ depends on unknown acquisition parameters, like exposure.
Hence recent and deep learning methods predict ``scale-invariant'' components, \ie, following $\I = \alpha \A \cdot \beta \S$.
This requires a subsequent (often manual) optimization for both $\alpha$ and $\beta$ \emph{individually}, which is prone to introduce distortion and inconsistencies.
It is argued that some intrinsic decomposition applications do not require absolute nor relative consistent $\A$ and $\S$~\cite{MITintrinsicImages}.

Conversely, we argue that energy preservation is important to allow for a subsequent principled and joint normalization of $\A$ and $\S$, \ie, tuning~$\alpha$ following $\I = \alpha \A \cdot \S/\alpha$.
We think $\A$ is defined by physical properties and should be considered an absolute value that is invariant to illumination changes and acquisition noise (\ie, illumination, sensor, etc).
The latter resides in $\S$, which could be further decomposed into subsequent contributions, but is out of the scope of this work.
Our definition is beneficial for applications relying on an illumination-invariant albedo, \eg, human face reconstruction.
So we target the respect of Eqn.~\eqref{eq:original_eq}.

In section~\ref{sec:intrinsicdecomposition:arch}, we present the CNN structure that, by construction, allows us to predict consistent $\A$ and $\S$ respecting Eqn.~\eqref{eq:original_eq}, including scale.
In section~\ref{sec:intrinsicdecomposition:GAN}, we present the loss function we optimize for, including a perceptually motivated one.

\subsection{Network architecture}
\label{sec:intrinsicdecomposition:arch}
Formally, we train a generative fully convolutional network $\N$ having parameters $\phi$ at predicting a coherent pair $(\Ap,\Sp)$ from an input image $\I$ such that
\begin{equation}
    \N(\I) = (\Ap ,\Sp)\,.
\end{equation}

\begin{figure}
    \begin{center}
        \subfloat{\includegraphics[width = \linewidth]{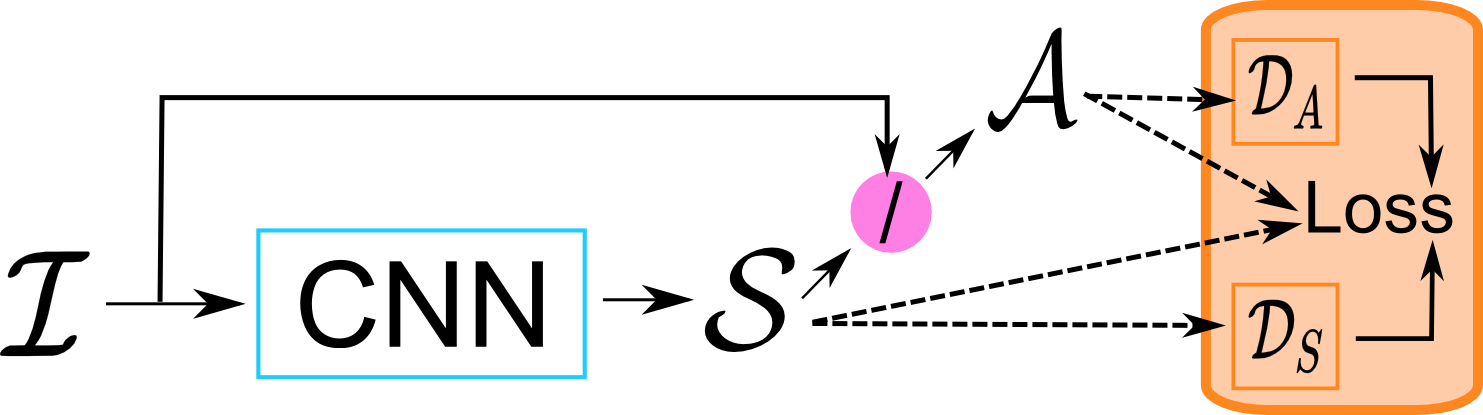}}
    \end{center}
    \vspace{-0.2cm}
    \caption{Global summary of the proposed GAN architecture.
    Element-wise division between shading and input (magenta) is used to define resulting albedo.
    Hence consistency is obtained by construction.
    The detailed architecture of the CNN (cyan box) is detailed in Fig~\ref{fig:localArch}.}
    \label{fig:globalArch}
    \vspace{-0.4cm}
\end{figure}

\paragraph{Consistent predictions.}
We impose the strict respect of Eqn.~\eqref{eq:original_eq} for our predictions: $\I = \Ap \cdot \Sp$.
So we predict either $\A$ or $\S$, and deduce the other by inverting the element-wise product of Eqn.~\eqref{eq:original_eq}.
This is naturally implemented in a CNN by incorporating an element-wise division of $\I$ by the predicted $\S$ at the end of the convolution layers, as illustrated in Fig.~\ref{fig:globalArch} by the magenta disc.
This does not hinder CNN training because both predicted elements can be used in the loss function and even brings the advantage that gradients are naturally fused and can safely be back-propagated.
Eqn.~\eqref{eq:original_eq} is respected \emph{by construction}, which we believe to be important for further processing in several applications.

In the remainder of this paper, we present our network as first estimating $\S$ and deducing $\A$ by division as shown in Fig.~\ref{fig:globalArch}.
Note that the inverse can be done without any additional effort: we compare both variants in the supplementary material.
Finally, we introduce the notations $\C$ (respectively $\Cp$), and the vocabulary ``component'', which generically represent either $\A$ or $\S$ (respectively $\Ap$ or $\Sp$).
We use it when the task is component-agnostic.

\paragraph{Residual learning.}
$\I$ and $\S$ (or $\A$) have a lot in common, hence the identity function is a good initial guess.
However, CNN are known to struggle when the solution is close to the identity~\cite{resnet}.
Residual networks have been proposed to tackle this drawback of deep networks.
We incorporate this structure into our network to facilitate the estimation of $\S$.

\begin{figure}
    \begin{center}
        \subfloat{\includegraphics[width = \linewidth]{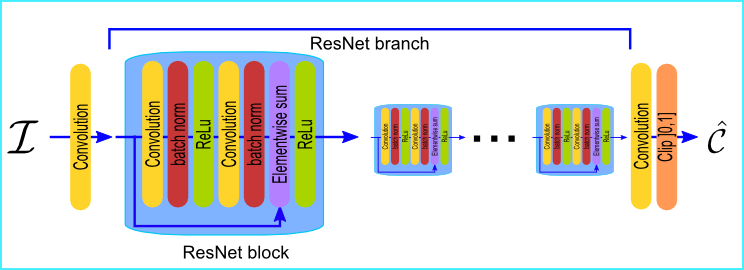}}
    \end{center}
    \vspace{-0.2cm}
    \caption{The generative CNN (cyan box in Fig.~\ref{fig:globalArch}) is composed of $10$ ResNet blocks in sequence.
    ResNet allows the CNN to focus on residual learning.
    }
    \label{fig:localArch}
    \vspace{-0.4cm}
\end{figure}

Fig.~\ref{fig:localArch} describes the CNN branch (cyan box in Fig.~\ref{fig:globalArch}) that learns the mapping from $\I$ to $\S$. We build it as a sequence of residual blocks as proposed in~\cite{resnet}.
A block is composed of two convolution layers of $64$ $3\times 3$ convolutions.
We use batch normalization~\cite{batchnorm} after every convolution as a regularization for the network. They are followed by a Rectified Linear unit (ReLu).
An element-wise summation with the input is done before the ReLu of the second convolution. $10$~such blocks are then connected in sequence.
Finally, note that our generative network is fully convolutional, hence it can process images of any size at test time.

\subsection{Loss function}
\label{sec:intrinsicdecomposition:GAN}
Let $\Igt$, $\Agt$ and $\Sgt$ denote the GT image, albedo and shading, respectively.
Loosely speaking, the goal of the learning task is to predict $\Ap$ and $\Sp$ ``as close as possible'' to $\Agt$ and $\Sgt$.
Formally, we train the network parameters $\phi$ to minimize a loss $L$ composed of three terms representing a data loss, a data gradient loss and an adversarial loss:
\begin{equation}
\label{eq:loss}
  \begin{split}
       L(\Ap,\Sp)= & L_{data}(\Ap,\Sp) + \Lgrad(\Ap,\Sp) 
       \\
       + & \lambda \Ladv(\Ap,\Sp)\,.
  \end{split}
\end{equation}

\paragraph{Data loss.}
First and foremost, data loss ensures that predictions fit the GT data.
As opposed to prior work~\cite{MITintrinsicImages,directintrinsic}, we want a measure sensitive to scale that imposes consistent global intensity.
Therefore, we define it as a $L_2$ norm over both the albedo and the shading:
\begin{equation}
    L_{data}(\Ap,\Sp) = ||\Agt - \Ap||^2 + ||\Sgt - \Sp||^2\,.
\end{equation}

\paragraph{Gradient loss.}
Both albedo and shading present sharp discontinuities which are essential, as well as smooth planar surfaces.
To favor estimations that exhibit variations consistent with these and avoid over-smoothing, we complement the loss with a $L_2$ norm over the gradient of both albedo and shading:
\begin{equation}
    \Lgrad(\Ap,\Sp)  = ||\nabla \Agt - \nabla \Ap||^2 
                     + ||\nabla \Sgt - \nabla \Sp||^2\,.
\end{equation}
We demonstrate its usefulness in the ablation study in the supplementary material.

\paragraph{Adversarial loss.}

\begin{figure}
\begin{center}
    \subfloat{\includegraphics[width = \linewidth]{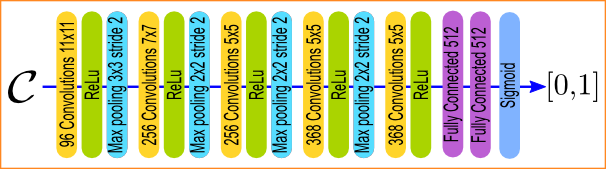}}
\end{center}
    \vspace{-0.2cm}
 \caption{The adversarial loss is defined by a continuously trained discriminator network. It is a binary classifier that differentiates between GT data and generated data predicted by the generative network (Fig.~\ref{fig:globalArch}).}
    \vspace{-0.4cm}
 \label{fig:adversarial}
\end{figure}

We observed that state of the art results show typical generative CNN visual artifacts.
Therefore we introduce an \emph{adversarial loss} for both $\A$ and $\S$.
This takes the form of two binary classifier CNN, called ``discriminators'', one per component: $\D_\A(\A)$ and $\D_\S(\S)$, respectively.
The discriminator's objective is to distinguish between the domain of generated versus GT data.
The goal of the generator network then becomes to fool the discriminator so that it cannot distinguish generated from GT images.
Hence, typical generative CNN artifacts are avoided.

Formally, we append the two discriminator networks $\D_\A$ and $\D_\S$ after both predictions $\Ap$ and $\Sp$, respectively (see Fig.~\ref{fig:globalArch}).
The whole forms a GAN~\cite{GAN}.
$\D_\A$ and $\D_\S$ are then used to form a (deep) perceptually-motivated loss layer for the generative network.
The loss it defined by

\begin{equation}
    L_{adv}(\Ap,\Sp)
    =
    - \log(
    \D_\A(\Ap)
    \cdot
    \D_\S(\Sp)
    )\,,
\end{equation}
which is low when $\D_\C$ cannot classify $\Cp$ to be a generated image.
The discriminator is fixed during a generator training iteration: it only serves as a perceptual measure.
Backpropagation through the discriminator enriches the error information by transforming it into a more expressive representation guiding the learning.
The drawback is that its intensity might surpass other error information coming from less deep locations, which requires the addition of a weighting parameter $\lambda$ in Eqn.~\eqref{eq:loss}.
The architecture of the discriminators follow standard classification architectures: convolutions with strided MaxPooling precede a few fully-connected layers predicting the probability (see Fig.~\ref{fig:adversarial}).

The discriminator is trained using a binary classification loss:
\begin{equation}
  \label{eq:discriminator_loss}
    L_{Discr}(\Cgt, \Cp) = -\log(\D_\C(\Cgt)) - \log(1 - \D_\C(\Cp))\,.
\end{equation}
We learn network parameters that maximize the likelihood for GT samples and minimize it for generated ones~\cite{GAN}.

\section{Training \& evaluation metrics}
\label{sec:training}
In this section, we discuss training data and details, as well as metrics used to quantitatively evaluate our results.

\subsection{Dataset}
Sintel is an open-source short computer animation movie that contains complex indoor and outdoor scenes.
For research purposes, it has been published in various formats, among which its intrinsic shading and albedo layers, as the ``MPI Sintel dataset''~\cite{SINTELdataset}.
It is available for 18 sequences, 17 of which are composed of 50 frames and one of 40, for a total of 890.
The albedo images have been rendered without illumination.
The shading images include illumination effects, and have been produced by rendering with a constant gray albedo.
Because of this creation process, the original frames from Sintel and the given corresponding GT do not respect Eqn.~\eqref{eq:original_eq}.
We recompose the original frames as $\I = \Agt \cdot \Sgt\,$ to obtain consistent data with respect to the GT.
This forms a consistent dataset called the ``ResynthSintel'' dataset used in our training and testing.

We use the two variants of training/testing splits as used in~\cite{directintrinsic}.
The \emph{scene split} assigns half of the movie's scenes to either set.
The \emph{image split} randomly assigns half of the frames to either set.
We use the latter for fair comparison to related work, but advise against its use: consecutive frames within a scene are very similar, so training on this split favors over-fitting networks.
The \emph{scene split} is more challenging and is the better evaluation: it requires more generalization capacity.

During training, we used standard data augmentation techniques by
randomly cropping patches of size $250 \times 250$ within the images after
scaling by a random factor in $[0.8, 1.2]$,
rotating by a random angle of maximum $15^{\circ}$,
and using random horizontal mirroring with a likelihood of~$0.5$.

\subsection{Evaluation}
For quantitative evaluation, error metrics have to be chosen.
To be able to compare to related work~\cite{Chen13,MITintrinsicImages,directintrinsic}, we consider the metrics used before: two data-related metrics (\siMSE and \siLMSE) and a perceptually motivated one (DSSIM).
However, because these metrics are insensitive to scale, we additionally consider scale-sensitive measures.
We now present all measures in detail, and will show that we outperform prior work on Sintel using all measures.

\subsubsection{Scale-Invariant Metrics}
\paragraph{\siMSE.}
Denoted MSE in previous work, we rename it \emph{scale invariant mean squared error} to avoid confusion:
\begin{equation}
\label{eq:simse}
	\text{\siMSE}(\Cp) = ||\Cgt-\alpha \Cp||^2/\nbpx \,,
\end{equation}
where $\C\in\{\A,\S\}$, $\alpha=\argmin{\alpha}||\Cgt-\alpha \Cp||^2$ and $\nbpx$ is the number of pixels in $\Cgt$.
Note that $\alpha$ is \emph{separately} optimized for $\Ap$ and $\Sp$.
Thus, any scale shift on either of them has no influence on the error, and errors in scale correlation between both predictions are not penalized either.
Despite the classical flaw of heavy outlier weighting~\cite{MSE}, \siMSE is a decent data-term, widely used in previous work and other applications like single-view depth estimation~\cite{EigenNIPS14}.

\paragraph{\siLMSE.}
Denoted LMSE in previous work, we rename it \emph{scale invariant local mean squared error}:
\begin{equation}
\text{\siLMSE} (\Cp) = \frac{1}{\nbpatches}\sum_{\Pp \in \Cp} \text{\siMSE}(\Pp) \,,
\end{equation}
where $\nbpatches$ is the number of patches and $\Pp$ is a square patch taken from $\Cp$ of size $10\%$ of its largest dimension.
Patches are regularly extracted on a grid so as to have a 50\% overlap between neighboring ones.
Note that in this case, the scale parameter $\alpha$ (see Eqn.~\eqref{eq:simse}) is optimized for each patch \emph{individually}.
This measure evaluates local structure similarity and is thus finer-grained than the MSE.

\paragraph{DSSIM.}
The \emph{structural similarity image index}~\cite{SSIM} is a perceptually-motivated measure that accounts for multiple independent structural and luminance differences.
It is here transformed into a dissimilarity measure:
\begin{equation}
\text{\DSSIM}(\Cp) = (1-\text{\SSIM}(\Cp))/2 \,.
\end{equation}

\begin{table*}[t]
    \begin{center}
        \begin{tabular}{|l||c|c|c||c|c|c||c|c|c|}
            \hline Sintel & \multicolumn{3}{|c||}{\siMSE} & \multicolumn{3}{|c||}{\siLMSE} & \multicolumn{3}{|c|}{DSSIM} \\
		    \emph{Image Split}              & A & S & Avg   & A & S & Avg   & A & S & Avg      \\
            \hline Baseline: Shading Constant               & 5.31 & 4.88 & 5.10 & 3.26 & 2.84 & 3.05 & 21.40 & 20.60 & 21.00 \\
            \hline Baseline: Albedo Constant                & 3.69 & 3.78 & 3.74 & 2.40 & 3.03 & 2.72 & 22.80 & 18.70 & 20.75 \\
            \hline Retinex \cite{MITintrinsicImages}	    & 6.06 & 7.27 & 6.67 & 3.66 & 4.19 & 3.93 & 22.70 & 24.00 & 23.35 \\
            \hline Lee et al. \cite{Lee2012}                & 4.63 & 5.07 & 4.85 & 2.24 & 1.92 & 2.08 & 19.90 & 17.70 & 18.80 \\
            \hline Barron et al. \cite{BarronTPAMI2015}     & 4.20 & 4.36 & 4.28 & 2.98 & 2.64 & 2.81 & 21.00 & 20.60 & 20.80 \\
            \hline Chen and Koltun \cite{Chen13}            & 3.07 & 2.77 & 2.92 & 1.85 & 1.90 & 1.88 & 19.60 & 16.50 & 18.05 \\
            \hline DirectIntrinsics \cite{directintrinsic} & \textbf{1.00} & \textbf{0.92} & \textbf{0.96} & 0.83 & 0.85 & 0.84 & 20.14 & 15.05 & 17.60 \\
            \hline Our work                                 & 1.24 & 1.28 & 1.26 & \textbf{0.69} & \textbf{0.70} & \textbf{0.70} & \textbf{12.63} & \textbf{12.13} & \textbf{12.38} \\
            \hline
        \end{tabular}
    \vspace{-0.2cm}
    \end{center}
    \caption{Quantitative scale-invariant results ($\times 100$) after double cross-validation on the Sintel \emph{image split}.}
    \label{tab:si_results}
    \vspace{-0.4cm}
\end{table*}

\subsubsection{Scale-Aware Metrics}
We argue that albedo has a unique physical value, and that the respect of Eqn.~\ref{eq:original_eq} is critical in many applications~\cite{BKPB17}.
Hence, the pair $(\A,\S)$ is unique.
Here we design metrics that take these observations into account:
we propose two data-terms measuring deviation from a consistent joint scale-dependent reconstruction.

\paragraph{\MSE.}
The traditional scale-sensitive MSE measures our overall goal: numerically get as close as possible to the GT:
\begin{equation}
\label{eq:MSE}
 \text{\MSE}(\Cp) = ||\Cgt-\Cp||^2/\nbpx \,.
\end{equation}

\paragraph{\rsMSE.}
Second, we introduce the \emph{relative scale mean squared error}.
As noted by~\cite{MITintrinsicImages}, Eqn.~\eqref{eq:MSE} is ``too strict for most algorithms on our data''.
While we acknowledge that the ill-posed nature of intrinsic decomposition leads to a scale ambiguity, we still think that Eqn.~\eqref{eq:original_eq} should be preserved.
Hence, we allow \emph{relative} tuning between $\Ap$ and $\Sp$, using a unique scale parameter~$\alpha$:
\begin{equation}
 \label{eq:RSMSE}
 \text{\rsMSE} (\Ap,\Sp)
 =
 \min_\alpha(||\Agt-\alpha \Ap||^2 + ||\Sgt- \Sp/\alpha||^2)/2\nbpx
  \,,
\end{equation}
This measures allows only relative scale optimization so that the following relationship is preserved:
\begin{equation}
 \label{eq:relative_scale_AS}
  \I = \alpha \A \cdot \S / \alpha \,,
\end{equation}
hence is consistent with Eqn.~\eqref{eq:original_eq}.
In other words: relative consistent intensity variations are tolerated with this measure, while global intensity as well as structural information must be preserved.

\subsection{Training details}

Our generative network is trained efficiently during 8K iterations with batch size $5$.
Since the first estimations of the generator network are poor, the adversarial training is not used during the $400$ first iterations to improve training stability~\cite{GAN}. 
Afterwards, the discriminators are trained alongside the generator in an iterative expectation-maximization-like procedure: we iterate between 3 discriminator updates with fixed generator, followed by 1 generator update with fixed discriminator, for a total of around $24K$ discriminator updates.
We empirically found that $\lambda=10^{-4}$ is the most efficient for our task.
We use the ADAM~\cite{adamopti} optimization method with a learning rate starting at $10^{-4}$ and decreasing to $10^{-6}$.
Complete training took around $20$ hours. 
Finally, we followed the double cross-validation procedure to produce our results, which are the average of two evaluation using networks trained on reciprocal testing and training sets~\cite{directintrinsic}.

\section{Results}
\label{sec:results}
In this section, we present qualitative and quantitative comparisons to related work.
First on the test data from the dataset we train on (section~\ref{sec:results:sintel}).
Second we study generalization to other data (section~\ref{sec:results:generalization}).
Note that in our supplementary material, we propose an ablation study measuring the impact of each component of our architecture.

\subsection{Training and evaluation on MPI Sintel}
\label{sec:results:sintel}

\begin{figure*}
    \begin{center}
        \subfloat{\includegraphics[width = \linewidth]{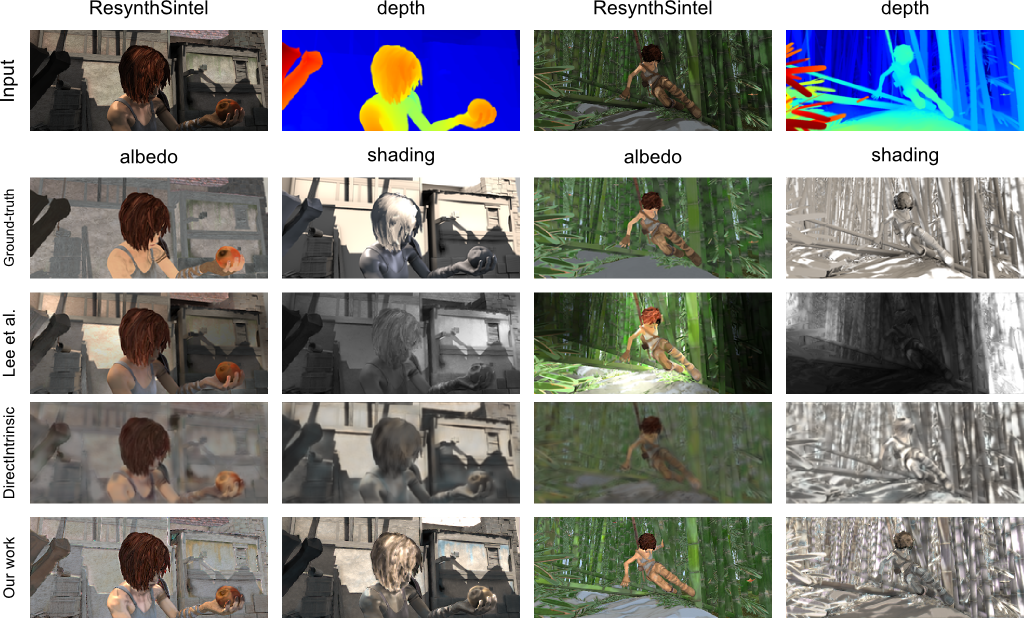}}\\

        \subfloat{\includegraphics[width = \linewidth]{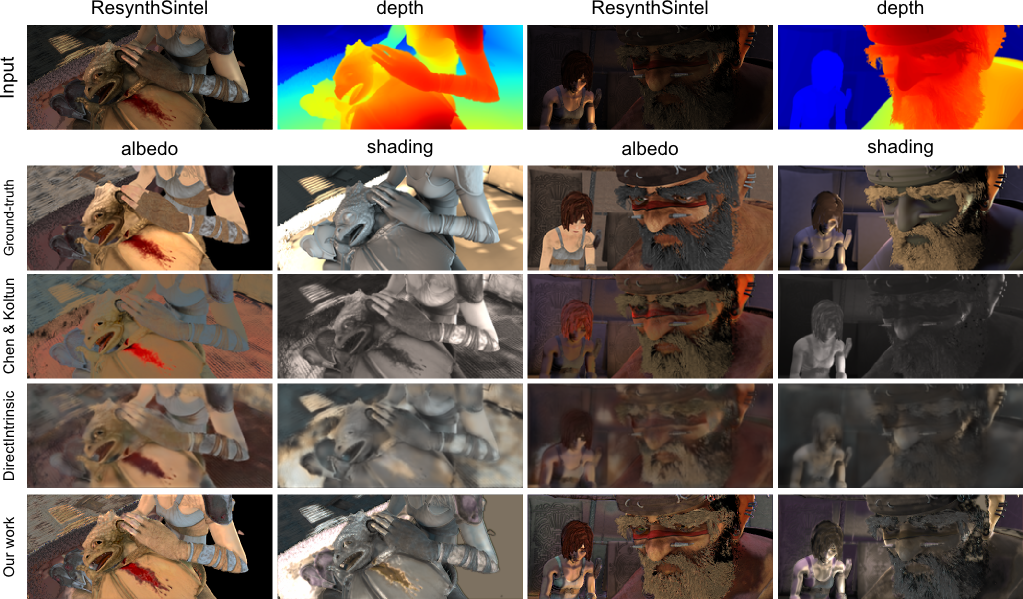}}
    \end{center}
    \caption{Qualitative results on the Sintel \emph{image split}. Comparison to~\cite{Lee2012,directintrinsic} (top) and~\cite{Chen13,directintrinsic} (bottom). Depth is not used by our method nor \cite{directintrinsic}.
    Our results are sharper thus closer to the GT, albeit emphasizing shading errors in the albedo images.
    }
    \label{fig:qualitative_cmp_1}
\end{figure*}

\begin{table*}[t]
    \begin{center}
     {
        \begin{tabular}{|l|c|c|c||c|c|c||c|c|c||c|c|c||c|}
            \hline 
            Sintel & \multicolumn{3}{|c||}{\siMSE} & \multicolumn{3}{|c||}{\siLMSE} & \multicolumn{3}{|c||}{\DSSIM} & \multicolumn{3}{|c||}{\MSE} & \rsMSE \\
            \emph{Scene Split}             & A & S & avg   & A & S & avg  & A & S & avg & A & S & Avg   &  \\
	    \hline \multicolumn{14}{|c|}{Evaluated on ResynthSintel} \\
            \hline \cite{directintrinsic}  & 2.09 & 2.76 & 2.42 & 1.08 & 1.31 & 1.19 & 22.75 & 19.37 & 21.06 & 2.50 & 4.83 & 3.66 & 3.09 \\
            \hline 
            Ours & \textbf{1.77} & \textbf{1.84} & \textbf{1.81} & \textbf{0.98} & \textbf{0.95} & \textbf{0.97} & \textbf{14.21} & \textbf{14.05} & \textbf{14.13} & \textbf{2.10} & \textbf{2.19} & \textbf{2.15} & \textbf{1.85} \\
            \hline \multicolumn{14}{|c|}{Reported in previous work} \\
            \hline \cite{directintrinsic}  & 2.01 & 2.24 & 2.13 & 1.31 & 1.48 & 1.39 & 20.73 & 15.94 & 18.33 & - & - & - & - \\
            \hline
        \end{tabular}
        }
    \end{center}
    \vspace{-0.2cm}
    \caption{Qualitative results ($\times 100$) on the Sintel \emph{scene split}. Top: comparison to~\cite{directintrinsic} using their published pre-trained network.
    Bottom: the reported values by~\cite{directintrinsic}, which were computed on a slightly differing Sintel dataset not respecting Eqn.~\eqref{eq:original_eq}.
    We included it for fair comparison, as it shows a lower bound that could be approached after tuning to ResynthSintel.}
    \label{tab:ss_results}
    \vspace{-0.4cm}
\end{table*}

We first compare on the scale-invariant metrics and the \emph{image split} test images (Tab.~\ref{tab:si_results} and Fig.~\ref{fig:qualitative_cmp_1}) for which many methods report results.
``DirectIntrinsics''~\cite{directintrinsic} (DI) and our work are the only deep learning approaches evaluating on dense GT data, and the only ones predicting solely from images, excluding depth.
Numbers show that the deep learning approaches substantially outperform classical methods.
A striking improvement of our method \wrt DI is the sharpness both in the spatial and color domain.
This can be seen on the albedo, where contours and especially the background is sharp.

Similarly, improved shading can be observed: the hair of the main character is well approximated (Fig.~\ref{fig:qualitative_cmp_1}, top left frame) thanks to the consistency enforced by Eqn.~\eqref{eq:original_eq}.
This is confirmed by the strong improvement on the \DSSIM measure, which penalizes local structural errors: our method corrects those by avoiding the typical generative CNN artifacts of blur, contrast oscillations and color bleeding.
This is also the cause of our improvement on the \siLMSE measure, because our results are locally more consistent.
Finally, we show slightly worse numerical results on the \siMSE metric on the \emph{image split} data (Tab.~\ref{tab:si_results}).
Recall that this does not consider scale nor albedo-shading consistency in its measure and that we optimize for the more challenging scale-sensitive recovery.

The most important numerical results are on the challenging \emph{scene split} case that requires better generalization capabilities to be handled well.
Tab.̣~\ref{tab:ss_results} shows our improvement on all numerical results.
The fact that the improvement is more significant on the scene split case hints that our method has better generalization capabilities than DI on the domain of Sintel-like rendered animation movies.
We refer the interested reader to our supplementary video for more results on the Sintel dataset:
the latter shows strong temporal consistency which is a beneficial outcome of having a fully convolutional (\ie, translation-invariant) architecture and of learning absolute values for albedo and shading.

\subsection{Inter-domain Generalization}
\label{sec:results:generalization}
Although we've shown that our trained model has an improved intra-domain generalization capacity, this is no guarantee for inter-domain generalizability.
Besides, Sintel is biased towards unrealistic appearances (\eg, blue shading).
To study generalization capacity, we apply typical training dataset tricks, \eg, mixing two datasets for training and evaluation of the trained model on test splits of both.

\paragraph{The MIT dataset~\protect\cite{MITintrinsicImages}} 
proposes realistic intrinsic images' decompositions into albedo and grayscale shading of 20 objects with 10 different illuminations.
Due to the small size and lack of variety, training a CNN on it is inappropriate.
Hence we mix it to Sintel and train on the union of both training splits: 445 frames of \emph{Sintel scene split}, and 100 images of \emph{MIT object split}~\cite{directintrinsic}.
We denote the trained model by ``\oursht''.
Tab.~\ref{tab:mixed_results} shows how a single such trained model evaluates on both test scenario's and compares to DI: our method outperforms previous work on both test cases, showing better adaptation capacity to larger image domains.

Balancing the mixing ratio in the dataset is also a widely used tuning trick: we tested sampling among both datasets with a likelihood of 50\%, and denote the corresponding trained model ``\oursff''.
In this case, our method's results logically improve even more on the MIT dataset while decreasing results on Sintel, albeit still improving over the state of the art methods on the stringent \MSE and \rsMSE metrics.
Such mixing can be further tuned at will with any newly available densely annotated dataset.

\begin{table}[t]
    \begin{center}
     {
        \begin{tabular}{|l|c|c|c|c|c|}
            \hline 
            & \small \siMSE & \small \siLMSE & \small\DSSIM & \small \MSE &  \small \rsMSE \\
            \hline \multicolumn{6}{|c|}{Evaluated on ResynthSintel \emph{scene split}} \\
            \hline \cite{directintrinsic}               & 2.42 &  \textbf{1.19} & 21.06 & 3.66 & 3.09 \\
            \hline \small \oursht                       & \textbf{2.41} & 1.24 & \textbf{16.65} & \textbf{2.80} & \textbf{2.45} \\
            \hline \small \oursff                       & 2.74 & 1.60 & 17.46 & 3.46 & 2.91 \\
            \hline \multicolumn{6}{|c|}{Evaluated on MIT \emph{object split}} \\    
            \hline \cite{directintrinsic}   & 2.24 & 0.91 & - & - & - \\
            \hline \small \oursht                       & 1.70 & 0.80 & 22.90 & 2.31 & 1.78\\
            \hline \small \oursff                       & \textbf{0.98} & \textbf{0.55} & \textbf{8.04} & \textbf{1.60} & \textbf{1.07}\\
            \hline
        \end{tabular}
        }
    \end{center}
    \vspace{-0.2cm}
    \caption{Quantitative results ($\times 100$) of our model trained on a mixed dataset and evaluated on the Sintel \emph{scene split} and MIT \emph{object split}.
    We compare to~\cite{directintrinsic}'s best networks on either evaluations (top: trained on Sintel + GenMIT, bottom: trained on their ResynthSintel + GenMIT).}
    \label{tab:mixed_results}
    \vspace{-0.4cm}
\end{table}

\paragraph{Intrinsic Images in the Wild} (IIW) is a large-scale realistic dataset, but only comes with sparse and relative GT annotations.
The WHDR metric compares to these~\cite{IntrinsicImageInWild14}.
It is thus not fit for training with our model, and it is hard to report quantitative results when testing our Sintel-trained one: like DI, we obtained a poor WHDR when evaluating on this dataset due to its sparse nature (which does not measure consistency of dense results).

Our method has not seen a real-world image at train time.
Nevertheless, $\Ap$ and $\Sp$ are decent on some examples while showing typical Sintel biases (gold and blue shading), as can be seen on the qualitative comparisons shown in Fig.~\ref{fig:IIW} and in supplementary material.
In general, it shows a lack of inter-domain generalization: our model is weakly suited to the irregularities of real data that do not exist in Sintel.

\paragraph{Discussion.}
Real-world data includes effects induced by optics and acquisition devices (\eg, white balancing), which impacts $\S$ in our model, its scale in particular.
Training on these variations requires an improved dense training dataset and additional experimentation studying suitability of scale-aware metrics in this setting.
We leave this for future work.

\begin{figure}
    \begin{center}
    {\setlength\tabcolsep{0mm}
    \begin{tabular}{c:cccccc}
      \subfloat{\includegraphics[width = 0.14\linewidth,trim={5cm 5cm 5cm 5cm},clip]{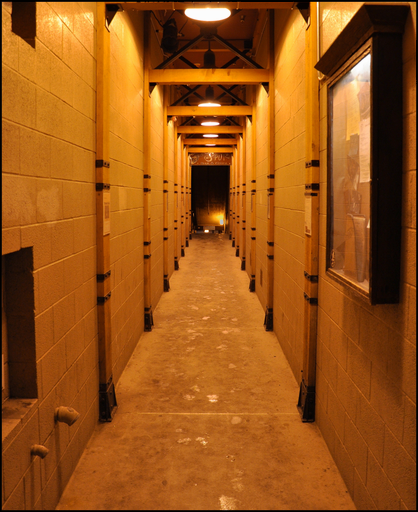}} &
      \subfloat{\includegraphics[width = 0.14\linewidth,trim={5cm 5cm 5cm 5cm},clip]{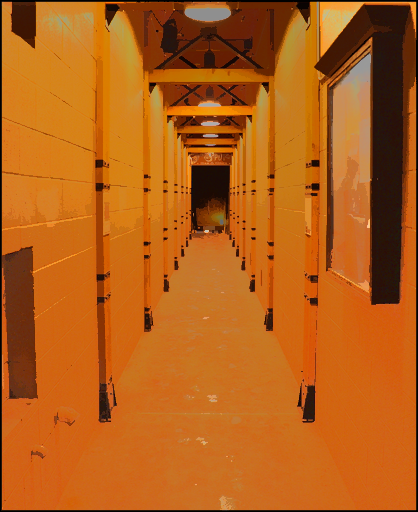}} &
      \subfloat{\includegraphics[width = 0.14\linewidth,trim={5cm 5cm 5cm 5cm},clip]{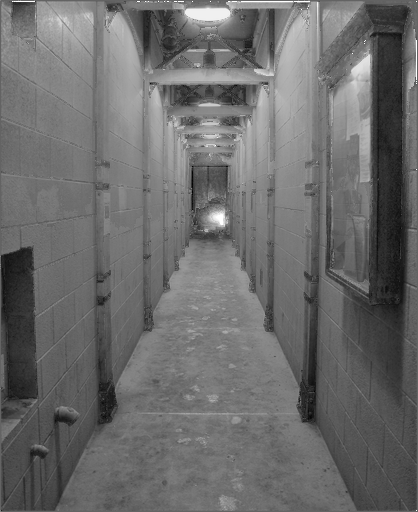}} &
      \subfloat{\includegraphics[width = 0.14\linewidth,trim={5cm 5cm 5cm 5cm},clip]{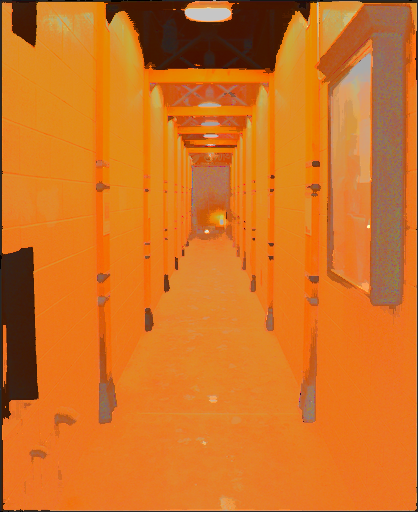}} &
      \subfloat{\includegraphics[width = 0.14\linewidth,trim={5cm 5cm 5cm 5cm},clip]{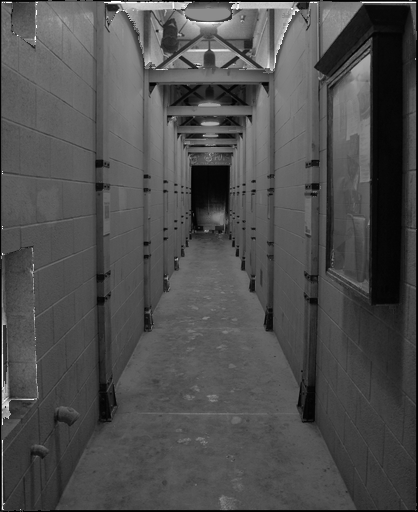}} & 
      \subfloat{\includegraphics[width = 0.14\linewidth,trim={5cm 5cm 5cm 5cm},clip]{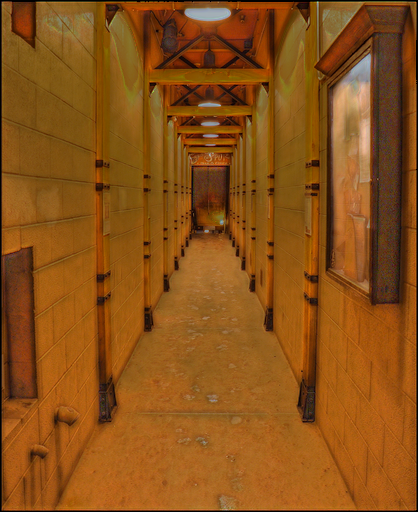}} &
      \subfloat{\includegraphics[width = 0.14\linewidth,trim={5cm 5cm 5cm 5cm},clip]{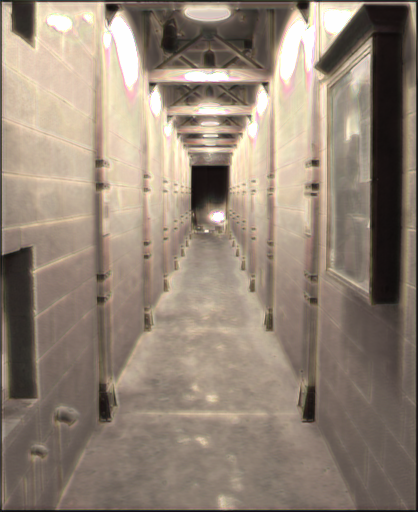}}
    \\
    $\I$ & \multicolumn{2}{c}{\cite{IntrinsicImageInWild14}} & \multicolumn{2}{c}{\cite{ZhouICCV15}} & \multicolumn{2}{c}{Ours}
    \\[-3mm]
        \end{tabular}
    }
    \end{center}
    \vspace{-0.2cm}
    \caption{Qualitative comparison of decomposing the $\I$ into $\Ap$ and $\Sp$, respectively, using̣~\cite{IntrinsicImageInWild14},~\cite{ZhouICCV15} and our method (from left to right).
    Our method decomposes the door better, but generates a yellowish shading.}
    \label{fig:IIW}
    \vspace{-0.4cm}
\end{figure}

\section{Conclusion}
\label{sec:conclu}
We presented a new architecture that learns single-image intrinsic decomposition and produces consistent energy-preserving results and scale-sensitive albedo.
We argue consistency makes it more usable for many applications, and therefore discussed and proposed metrics to measure it. 
Our contribution outperforms state-of-the-art methods on dense decomposition benchmarks, and presents better intra-domain generalization capabilities.
By mixing training datasets, generalization is possible for small domain shifts, but this excludes real-world data. 
This limitation is a great inspiration for future work as currently the only existing GT is unrealistic (Sintel), small (MIT) or sparse (IIW).
Exploiting the power of deep learning for realistic decomposition therefore requires domain shift techniques and/or directly tackling the problem of dense intrinsic decomposition of real data.

{\small
\bibliographystyle{ieee}
\bibliography{egbib}
}

\end{document}